\documentclass{article}
\usepackage{times}
\usepackage{caption}
\PassOptionsToPackage{numbers, compress}{natbib}
\usepackage[final]{nips_2016}

\usepackage[utf8]{inputenc} % allow utf-8 input
\usepackage[T1]{fontenc}    % use 8-bit T1 fonts
\usepackage{hyperref}       % hyperlinks
\usepackage{url}            % simple URL typesetting
\usepackage{booktabs}       % professional-quality tables
\usepackage{amsfonts}       % blackboard math symbols
\usepackage{nicefrac}       % compact symbols for 1/2, etc.
\usepackage{microtype}      % microtypography

\usepackage{url}
\usepackage{latexsym}
\usepackage{hyperref}

\usepackage{algorithm}
\usepackage{algorithmic}
\usepackage[scientific-notation=true]{siunitx}
\usepackage{amsmath}
\usepackage{amsthm}
\usepackage{amssymb}
\usepackage{graphicx}
\usepackage{xspace}
\usepackage{tabularx}
\usepackage{multicol}
\usepackage{multirow}
\usepackage{url}
\usepackage{wrapfig}
\usepackage{comment}
\usepackage{listings}
\usepackage{color}
\usepackage[utf8]{inputenc}
\usepackage{fancyvrb}
\usepackage{booktabs}
\usepackage{color}
\usepackage[normalem]{ulem}
\usepackage{booktabs}

\newcommand{\bmx}[0]{\begin{bmatrix}}
\newcommand{\emx}[0]{\end{bmatrix}}

\newcommand{\vect}[1]{\mathbf{#1}}
\newcommand{\vects}[1]{\boldsymbol{#1}}

\newcommand{\vb}[0]{\vect{b}}

\newcommand{\vn}[0]{\vect{n}}
\newcommand{\vh}[0]{\vect{h}}

\newcommand{\vx}[0]{\vect{x}}

\newcommand{\PP}[0]{\vects{\phi}}
\newcommand{\PS}[0]{\vects{\psi}}

\newcommand{\vmu}[0]{\vects{\mu}}
\newcommand{\vepsilon}[0]{\vects{\epsilon}}

\newcommand{\LL}[0]{\mathcal{L}}

\newcommand{\Bernoulli}{\text{Bernoulli}}
\newcommand{\E}[0]{\mathbb{E}}

\title{Iterative Refinement of the Approximate Posterior for Directed Belief Networks}

\author{
R Devon Hjelm \\
University of New Mexico and the Mind Research Network \\
 \texttt{dhjelm@mrn.org} \\
 \And 
 Kyunghyun Cho\\
 Courant Institute \& Center for Data Science, New York University \\
\texttt{kyunghyun.cho@nyu.edu} \\
 \And
 Junyoung Chung\\
 University of Montreal \\
 \texttt{junyoung.chung@umontreal.ca}\\
 \And
 Russ Salakhutdinov\\
 Carnegie Melon University\\
 \texttt{rsalakhu@cs.toronto.edu}\\
 \And
 Vince Calhoun\\
 University of New Mexico and the Mind Research Network\\
 \texttt{vcalhoun@mrn.org}
 \And
 Nebojsa Jojic\\
 Microsoft Research\\
\texttt{jojic@microsoft.com}
}

\begin{document}

\maketitle

\begin{abstract}
Variational methods that rely on a recognition network to approximate the
posterior of directed graphical models offer better inference and learning than
previous methods.  Recent advances that exploit the capacity and flexibility in
this approach have expanded what kinds of models can be trained.  However, as a proposal for the
posterior, the
capacity of the recognition network is limited, which can constrain the representational power of the generative
model and increase the variance of Monte Carlo estimates. To address these
issues, we introduce an iterative refinement procedure for improving the
approximate posterior of the recognition network and show that training with the
refined posterior is competitive with state-of-the-art methods.  The advantages
of refinement are further evident in an increased effective sample size, which
implies a lower variance of gradient estimates.
\end{abstract}

\section{Introduction}
Variational methods have surpassed traditional methods such as Markov chain
Monte Carlo \citep[MCMC,][]{neal1992connectionist} and mean-field coordinate
ascent \citep{saul1996mean} as the de-facto standard approach for training
directed graphical models.  Helmholtz machines~\citep{dayan1995helmholtz} are a
type of directed graphical model that 
%succeed primarily because they 
approximate the posterior distribution with a \emph{recognition network} that
provides fast inference as well as flexible learning which scales well to large
datasets.  Many recent significant advances in training Helmholtz machines come
as estimators for the gradient of the objective w.r.t. the approximate
posterior.  The most successful of these methods, variational autoencoders
\citep[VAE,][]{kingma2013auto}, relies on a re-parameterization of the latent
variables to pass the learning signal to the recognition network.  This type of
parameterization, however, is not available with discrete units, and the naive
Monte Carlo estimate of the gradient has too high variance to be practical
\citep{dayan1995helmholtz,kingma2013auto}.

However, good estimators are available through importance sampling
\citep{bornschein2014reweighted}, input-dependent baselines
\citep{mnih2014neural}, a combination baselines and importance sampling
\citep{mnih2016variational}, and parametric Taylor expansions
\citep{gu2015muprop}. Each of these methods strive to be a lower-variance
and unbiased gradient estimator.  However, the reliance on the recognition
network means that the quality of learning  is bounded by the capacity of the
recognition network, which in turn raises the variance.  

We demonstrate reducing
the variance of Monte Carlo based estimators by iteratively refining the
approximate posterior provided by the recognition network. 
The complete learning algorithm follows
expectation-maximization~\citep[EM,][]{dempster1977maximum, neal1998view},
where in the E-step the variational parameters of the approximate posterior are
initialized using the recognition network, then iteratively refined.  The
refinement procedure provides an asymptotically-unbiased estimate of the
variational lowerbound, which is tight w.r.t. the true posterior and can be
used to easily train both the recognition network and generative model during
the M-step.  The variance-reducing refinement is available to any directed
graphical model and can give a more accurate estimate of the log-likelihood of
the model.  

For the iterative refinement step, we use adaptive importance
sampling~\citep[AIS,][]{oh1992adaptive}.  We demonstrate the proposed refinement
procedure is effective for training directed belief networks, providing a better
or competitive estimates of the log-likelihood.  We also demonstrate the
improved posterior from refinement can improve inference and
accuracy of evaluation for models trained by other methods.

\section{Directed Belief Networks and Variational Inference}
A \emph{directed belief network} is a generative directed graphical model
consisting of a conditional density $p(\vx | \vh)$ and a prior $p(\vh)$, such
that the joint density can be expressed as $p(\vx, \vh) = p(\vx | \vh) p(\vh)$.
In particular, the joint density factorizes into a hierarchy of
conditional densities and a prior: $p(\vx , \vh) = p(\vx | \vh_1) p(\vh_L) \prod_{l=1}^{L-1} p(\vh_l
| \vh_{l+1})$, where $p(\vh_l | \vh_{l + 1})$ is the conditional density at the
$l$-th layer and $p(\vh_L)$ is a prior distribution of the top layer.  Sampling
from the model can be done simply via ancestral-sampling, first sampling from
the prior, then subsequently sampling from each layer until reaching the
observation, $\vx$.  This latent variable structure can improve model capacity,
but inference can still be intractable, as is the case in sigmoid belief
networks \citep[SBN,][]{neal1992connectionist}, deep belief networks
\citep[DBN,][]{hinton2006fast}, deep autoregressive networks \citep[DARN,
][]{gregor2013deep}, and other models in which each of the conditional
distributions involves complex nonlinear functions.

\subsection{Variational Lowerbound of Directed Belief Network}
The objective we consider is the likelihood function, $p(\vx;\PP)$, where $\PP$
represent parameters of the generative model (e.g. a directed belief network).
Estimating the likelihood function given the joint distribution, $p(\vx,
\vh;\PP)$, above is not generally possible as it requires intractable
marginalization over $\vh$.  Instead, we introduce an approximate posterior,
$q(\vh | \vx)$, as a proposal distribution.  In this case, the log-likelihood
can be bounded from below\footnote{
    For clarity of presentation, we will often omit dependence on parameters
    $\PP$ of the generative model, so that $p(\vx, \vh) = p(\vx, \vh;\PP)$
}:
\begin{align}
    \label{eq:approx_logp}
    \log p(\vx) 
    = \sum_{\vh} \log p(\vx, \vh) 
    \geq \sum_{\vh} q(\vh | \vx) \log \frac{p(\vx, \vh)}{q(\vh | \vx)} 
    = \mathbb{E}_{q(\vh | \vx)}\left[\log \frac{p(\vx, \vh)}{q(\vh | \vx)} \right] 
    := \LL_1,
\end{align}
where we introduce the subscript in the lowerbound to make the connection to
importance sampling later.  The bound is tight (e.g., $\LL_1 = \log p(\vx)$)
when the KL divergence between the approximate and true posterior is zero (e.g.,
$D_{KL}(q(\vh | \vx) || p(\vh | \vx)) = 0$).  The gradients of the lowerbound
w.r.t. the generative model can be approximated using the Monte Carlo
approximation of the expectation: 
\begin{align} \label{eq:grad1}
\nabla_{\PP} \LL_1 \approx \frac{1}{K} \sum_{k=1}^K \nabla_{\PP} \log p(\vx, \vh^{(k)}; \PP),
\hspace{0.1in} \vh^{(k)} \sim q(\vh | \vx).
\end{align}

The success of variational inference lies on the choice of approximate
posterior, as poor choice can result in a looser variational bound.  A deep
feed-forward \emph{recognition network} parameterized by $\PS$ has become a
popular choice, such that $q(\vh | \vx) = q(\vh | \vx; \PS)$, as it offers fast
and flexible data-dependent inference \citep[see,
e.g.,][]{salakhutdinov2010efficient, kingma2013auto, mnih2014neural,
rezende2014stochastic}.  Generally known as a ``Helmholtz machine''
\citep{dayan1995helmholtz}, these approaches often require additional tricks to
train, as the naive Monte Carlo gradient of the lowerbound w.r.t. the
variational parameters has high variance.  In addition, the variational
lowerbound in Eq.~\eqref{eq:approx_logp} is constrained by the assumptions
implicit in the choice of approximate posterior, as the approximate posterior
must be within the capacity of the recognition network and factorial.

\subsection{Importance Sampled Variational lowerbound}
These assumptions can be relaxed by using an unbiased $K$-sampled importance
weighted estimate of the likelihood function (see \citep{burda2015importance}
for details): 
\begin{align}
\label{eq:isest}
\LL_1 \leq \LL_K = \frac{1}{K} \sum_{k=1} \frac{p(\vx, \vh^{(k)})}{q(\vh^{(k)} | \vx)} = \frac{1}{K} \sum_{k=1} w^{(k)} \leq p(\vx),
\end{align}
where $\vh^{(k)} \sim q(\vh | \vx)$ and $w^{(k)}$ are the importance weights.
This lowerbound is tighter than the single-sample version provided in
Eq.~\eqref{eq:approx_logp} and is an asymptotically unbiased estimate of the
likelihood as $K \rightarrow \infty$.

The gradient of the lowerbound w.r.t. the model parameters $\PP$ 
is simple and can be estimated as:
\begin{align} \label{eq:gradK}
\nabla_{\PP} \LL_K =  \sum_{k=1}^K \tilde{w}^{(k)} \nabla_{\PP} \log p(\vx, \vh^{(k)}; \PP), 
\quad \text{where} \ \tilde{w}^{(k)} = \frac{w^{(k)}}{\sum_{k'=1}^K w^{(k')}}.
\end{align}

The estimator in Eq.~\eqref{eq:isest} can reduce the variance
of the gradients, $\nabla_{\PS} \LL_K$, but in general additional variance
reduction is needed \citep{mnih2016variational}.  Alternatively, importance
sampling yields an estimate of the inclusive KL divergence, $D_{KL}(p(\vh | \vx)
|| q(\vh | \vx))$, which can be used for training parameters $\PS$ of the
recognition network \citep{bornschein2014reweighted}.  However, it is well known
that importance sampling can yield heavily-skewed distributions over the
importance weights \citep{doucet2001introduction}, so that only a small number
of the samples will effectively have non-zero weight.  This is consequential not
only in training, but also for evaluating models when  using Eq.~\eqref{eq:isest} to estimate test log-probabilities, which
requires drawing a very large number of samples ($N \geq 100,000$ in the
literature for models trained on MNIST \citep{gregor2013deep}).

The effective samples size, $\vn_e$, of importance-weighted estimates increases
and is optimal when the approximate posterior matches the true posterior:
\begin{align}
    \label{eq:ess}
\vn_e = \frac{\left(\sum_{k=1}^K w^{(k)}\right)^2}{\sum_{k=1}^K (w^{(k)})^2}
\leq \frac{\left(\sum_{k=1}^K p(\vx, \vh^{(k)}) / p(\vh^{(k)} | \vx) \right)^2}{\sum_{k=1}^K \left( p(\vx, \vh^{(k)}) / p(\vh^{(k)} | \vx)\right)^2} 
\leq \frac{\left(K p(\vx)\right)^2}{K p(\vx)^2} 
= K.
\end{align}
Conversely, importance sampling from a poorer approximate posterior will have
lower effective sampling size, resulting in higher variance of the gradient
estimates.  In order to improve the effectiveness of importance sampling, we
need a method for improving the approximate posterior from those provided by the
recognition network.

\section{Iterative Refinement for Variational Inference (IRVI)}

\begin{figure}[]
\begin{center}
\includegraphics[scale=.4]{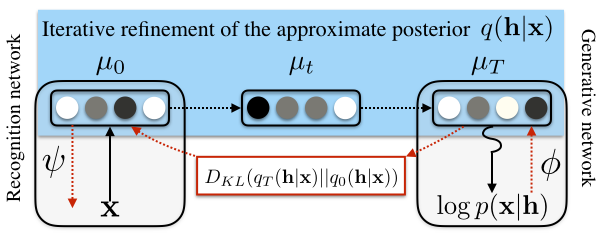}
\end{center}
\caption{ \small
Iterative refinement for variational inference.  An initial estimate of the
variational parameters is made through a recognition network.  The variational
parameters are then updated iteratively, maximizing the lowerbound.  The final
approximate posterior is used to train the generative model by sampling.  The
recognition network parameters are updated using the KL divergence between the
refined posterior $q_k$ and the output of the recognition network $q_0$.}
\label{fig:IRVI}
\end{figure}

To address the above issues, iterative refinement for variational inference
(IRVI) uses the recognition network as a preliminary guess of the posterior,
then refines the posterior through iterative updates of the variational
parameters.  For the refinement step, IRVI uses a stochastic transition
operator, $g(.)$, that maximizes the variational lowerbound.

An overview of IRVI is available in Figure \ref{fig:IRVI}.  For the expectation
(E)-step, we feed the observation~$\vx$ through the recognition network to get
the initial parameters, $\vmu_0$, of the approximate posterior, $q_0(\vh | \vx;
\PS)$.  We then refine $\vmu_0$ by applying $T$ updates to the variational
parameters, $\vmu_{t+1} = g(\vmu_{t}, \vx)$, iterating through $T$
parameterizations $\vmu_1, \dots, \vmu_T$ of the approximate posterior $q_t (\vh
| \vx)$. 

With the final set of parameters, $\vmu_T$, the gradient estimate of the
recognition parameters $\PS$ in the maximization (M)-step is taken w.r.t the
negative exclusive KL divergence:
\begin{align}
-\nabla_{\PS} D_{KL}(q_T(\vh | \vx) || q_0(\vh | \vx; \PS)) 
\approx  \frac{1}{K} \sum_{k=1}^K \nabla_{\PS} \log q_0(\vh^{(k)} | \vx; \PS), 
\label{eq:qterm} 
\end{align}
where $\vh^{(k)} \sim q_T(\vh | \vx)$.  Similarly, the gradients w.r.t. the
parameters of the generative model $\PP$ follow Eqs.~\eqref{eq:grad1} or
\eqref{eq:gradK} using samples from the refined posterior $q_T(\vh | \vx)$.  As
an alternative to Eq.~\eqref{eq:qterm}, we can maximize the negative inclusive
KL divergence using the refined approximate posterior:
\begin{align}
\label{eq:qterm2}
-\nabla_{\PS} D_{KL}(p(\vh | \vx) || q_0(\vh | \vx; \PS)) 
\approx  \sum_{k=1}^K \tilde{w}^{(k)} \nabla_{\PS} \log q_0(\vh^{(k)} | \vx; \PS).
\end{align}

The form of the IRVI transition operator, $g(\vmu_{t}, \vx)$, depends on the
problem.  In the case of continuous variables, we can make use of the VAE
re-parameterization with the gradient of the lowerbound in
Eq.~\eqref{eq:approx_logp} for our refinement step (see supplementary material).
However, as this is not available with discrete units, we take a different
approach that relies on adaptive importance sampling.
%Eq.~\eqref{eq:rec_train} is central to training with IRVI: we train the recognition network to \emph{predict} the refined approximate posterior,  improving the initialization for future inference and avoiding the need to back-propagate gradients through the latent variables.

%It is easy to demonstrate such a transition operator, $g(\TT, \vx)$, exists. 
%With variational autoencoders (VAE), the back-propagated gradient of the lowerbound with respect to the approximate posterior is composed of individual gradients for each factor, $\theta_i$ that can be applied simultaneously. 
%Applying the gradient directly to the variational parameters, $\TT$, without back-propagating to the recognition network parameters, $\PS$, yields a simple iterative refinement operator:
%\begin{align}
%\label{eq:GDIR}
%\TT_{t + 1} = g(\TT_t, \vx, \gamma) = \TT_t + \gamma \nabla_{\TT} \LL(\TT, \vx, \vepsilon),
%\end{align}
%where $\gamma$ is the inference rate hyperparameter and $\vepsilon$ is auxiliary noise used in the re-parameterization.

%This gradient-descent iterative refinement (GDIR) is very straightforward with continuous latent variables as with VAE.
%However, GDIR with discrete units suffers the same shortcomings as when passing the gradients directly,  so a better transition operator is needed.

\subsection{Adaptive Importance Refinement (AIR)}
\emph{Adaptive importance sampling} \citep[AIS,][]{oh1992adaptive} provides a
general approach for iteratively refining the variational parameters.  For
Bernoulli distributions, we observe that the mean parameter of the true
posterior, $\hat{\vmu}$, can be written as the expected value of the latent
variables:
\begin{align}
\label{eq:AIS}
\hat{\vmu} = \E_{p(\vh | \vx)} \left[ \vh \right] 
= \sum_{\vh} \vh \ p(\vh | \vx) 
= \frac{1}{p(\vx)} \sum_{\vh} q(\vh | \vx) \ \vh \ \frac{p(\vx, \vh)}{q(\vh | \vx)} 
\approx \sum_{k=1}^K \tilde{w}^{(k)} \vh^{(k)}.
\end{align}

%For binary stochastic variables with the Bernoulli centers, $\vmu = \TT$, for each update or \emph{inference step}, $k$, we draw $M$ samples from the Bernoulli distribution, then move the centers using the importance weights:
%The AIS update is an asymptotically unbiased estimate of the center of the posterior.
As the initial estimator typically has high variance, AIS iteratively moves
$\vmu_t$ toward $\hat{\vmu}$ by applying Eq.~\ref{eq:AIS} until a stopping
criteria is met.  While using the update, $g(\vmu_t, \vx, \gamma) = \sum_{k=1}^K
\tilde{w}^{(k)} \vh^{(k)}$ in principle works, a convex combination of
importance sample estimate of the current step and the parameters from the
previous step tends to be more stable:
\begin{align}
    \vh^{(m)} \sim \Bernoulli(\vmu_k); \quad
\vmu_{t+1} = g(\vmu_t, \vx, \gamma) = (1 - \gamma) \vmu_t + \gamma \sum_{k=1}^K \tilde{w}^{(k)} \vh^{(k)}.
\end{align}
Here, $\gamma$ is the inference rate and $(1 - \gamma)$ can be thought of as the
adaptive ``damping'' rate. 
%As required for IRVI, the true posterior is a stationary state, as the unnormalized importance weights become $p(\vx, \vh^{(m)}) / p(\vh^{(m)} | \vx) = p(\vx)$, making the weights uniform across samples, so that $g(\vmu_k, \vx) = (1 - \gamma) \vmu_k + \gamma \vmu_k = \vmu_k$. 
%This is an asymptotically unbiased estimate of the lowerbound that works very well in practice with a finite number of samples. 
This approach, which we call adaptive importance refinement (AIR), should work
with any discrete parametric distribution.  Although AIR is applicable with
continuous Gaussian variables, which model second-order statistics, we leave
adapting AIR to continuous latent variables for future work.

\subsection{Algorithm and Complexity}
The general AIR algorithm follows Algorithm~\ref{al:IRVI} with gradient
variations following Eqs.~\eqref{eq:grad1}, \eqref{eq:gradK}, \eqref{eq:qterm},
and \eqref{eq:qterm2}.  While iterative refinement may reduce the variance of
stochastic gradient estimates and speed up learning, it comes at a computational
cost, as each update is $T$ times more expensive than fixed approximations.
However, in addition to potential learning benefits, AIR can also improve the
approximate posterior of an already trained directed belief networks at test,
independent on how the model was trained. Our implementation following Algorithm
\ref{al:IRVI} is available at \href{https://github.com/rdevon/IRVI}{https://github.com/rdevon/IRVI}.

\begin{algorithm}                      
\caption{AIR}
\label{al:IRVI} 
\small
\begin{algorithmic}
\REQUIRE A generative model $p(\vx, \vh; \PP) = p(\vx | \vh; \PP) p(\vh;\PP)$ and a recognition network $\vmu_0 = f(\vx; \PS)$
%\REQUIRE Number of iterations, $K$
%\REQUIRE (For AIR) Number of adaptive samples, $M$
\REQUIRE A transition operator $g(\mu, \vx, \gamma)$ and inference rate $\gamma$.
%\STATE (E-step for training or test)
\STATE Compute $\vmu_0 = f(\vx; \PS)$ for $q_0(\vh | \vx; \PS)$
\FOR{t=1:T}
\item Draw $K$ samples $\vh^{(k)} \sim q_t(\vh | \vx)$ and compute normalized importance weights $\tilde{w}^{(k)}$
\item $\vmu_t = (1 - \gamma) \vmu_{t-1} + \gamma \sum_{k=1}^K \tilde{w}^{(k)} \vh^{(k)}$  
\ENDFOR
\IF{reweight}
\STATE $\Delta \PP \propto \sum_{k=1}^K \tilde{w}^{(k)} \nabla_{\PP} \log p(\vx, \vh^{(k)}; \PP)$
\ELSE
\STATE $\Delta \PP \propto \frac{1}{K} \sum_{k=1}^K \nabla_{\PP} \log p(\vx, \vh^{(k)}; \PP)$
\ENDIF
\IF{inclusive KL Divergence}
\STATE $\Delta \PS \propto \sum_{k=1}^K \tilde{w}^{(k)} \nabla_{\PS} \log q_0(\vh^{(k)} | \vx; \PS)$
\ELSE
\STATE $\Delta \PS \propto \frac{1}{K} \sum_{k=1}^K \nabla_{\PS} \log q_0(\vh^{(k)} | \vx; \PS)$
\ENDIF
\end{algorithmic}
\end{algorithm}

\section{Related Work}

Adaptive importance refinement (AIR) trades computation for expressiveness and
is similar in this regard to the refinement procedure of hybrid MCMC for
variational inference \citep[HVI,][]{icml2015_salimans15} and normalizing flows
for VAE \citep[NF,][]{rezende2015variational}.  HVI has a similar complexity as
AIR, as it requires re-estimating the lowerbound at every step.  While NF can
be less expensive than AIR, both HVI and NF rely on the VAE re-parameterization
to work, and thus cannot be applied to discrete variables.  Sequential
importance sampling~\citep[SIS,][]{doucet2001introduction} can offer a better refinement
step than AIS but typically requires resampling to control variance.  While
parametric versions exist that could be applicable to training directed
graphical models with discrete units \citep{gu2015neural, paige2016inference},
their applicability as a general refinement procedure is limited as the
refinement parameters need to be learned.

%AIR also shares similarities with 
Importance sampling is central to reweighted
wake-sleep~\citep[RWS,][]{bornschein2014reweighted}, importance-weighted
autoencoders~\citep[IWAE,][]{burda2015importance}, variational inference for
Monte Carlo objectives~\citep[VIMCO,][]{mnih2016variational}, and recent work on
stochastic feed-forward networks \citep[SFFN,][]{tang2013learning,
raiko2014techniques}. 
%Iterative refinement, however, takes a distinctly orthogonal approach, using importance sampling to achieve a better posterior through refinement. 
%While the end result may be a reduction of variance during training, none of these methods can refine the posterior further at test.
While each of these methods are competitive, they rely on importance samples
from the recognition network and do not offer the low-variance estimates
available from AIR.  Neural variational inference and
learning~\citep[NVIL,][]{mnih2014neural} is a single-sample and biased version
of VIMCO, which is greatly outperformed by techniques that use importance
sampling. Both NVIL and VIMCO reduce the variance of the Monte Carlo estimates of gradients
by using an input-dependent baseline, but this approach does not necessarily provide
a better posterior and cannot be used to give better estimates of the likelihood
function or expectations.

%re-weighted wake-sleep \citep[RWS,][]{bornschein2014reweighted}, like adaptive importance sampling iterative refinement (AIR), offer arguably the best solution for inference and learning in directed belief networks with discrete variables, though NVIL is biased, and RWS (like AIR) is asymptotically so. 
%While NVIL generally works, the variance is still not low enough to be practical, and convergence takes \emph{much} longer in terms of epochs and wall clock time than AIR, despite lower complexity. 
%RWS is very successful at training SBNs, but its complex objective function means that convergence properties cannot be proven. 

Finally, IRVI is meant to be a general approach to refining the approximate
posterior.  IRVI is not limited to the refinement step provided by AIR, and many
different types of refinement steps are available to improve the posterior for
models above (see supplementary material for the continuous case).  SIS and sequential
importance resampling~\citep[SIR,][]{gordon1993novel} can be used as an alternative to AIR and may provide a
better refinement step for IRVI.

\section{Experiments}
We evaluate iterative refinement for variational inference (IRVI) using adaptive
importance refinement (AIR) for both training and evaluating directed belief
networks. We train and test on the following benchmarks: the binarized MNIST
handwritten digit dataset \cite{salakhutdinov2008quantitative} and the
Caltech-101 Silhouettes dataset. We centered the MNIST and Caltech datasets by
subtracting the mean-image over the training set when used as input to the
recognition network. We also train additional models using the re-weighted
wake-sleep algorithm \citep[RWS,][]{bornschein2014reweighted}, the
state of the art for many configurations of directed belief networks with
discrete variables on these datasets for comparison and to demonstrate improving
the approximate posteriors with refinement.  With our experiments, we show that
1) IRVI can train a variety of directed models as well or better
than existing methods, 2) the gains from refinement improves the approximate
posterior, and can be applied to models trained by other algorithms, and 3)
IRVI can be used to improve a model with a relatively simple
approximate posterior.

Models were trained using the RMSprop algorithm~\citep{Hinton-Coursera2012} with
a batch size of $100$ and early stopping by recorded best variational lower
bound on the validation dataset. For AIR, $20$ ``inference steps" ($K=20$), $20$
adaptive samples ($M=20$), and an adaptive damping rate, $(1 - \gamma)$, of $0.9$ were
used during inference, chosen from validation in initial experiments. $20$
posterior samples ($N=20$) were used for model parameter updates for both AIR
and RWS. All models were trained for $500$ epochs and were fine-tuned for an
additional $500$ with a decaying learning rate and SGD.

We use a generative model composed of a) a factorized Bernoulli prior as with
sigmoid belief networks (SBNs) or b) an autoregressive prior, as in published
MNIST results with deep autoregressive networks \citep[DARN,
][]{gregor2013deep}:
\begin{align}
\text{a)} \ p(\mathbf{h}) = \prod_i p(h_i); P(h_i=1) = \sigma(b_i), \quad
\text{b)} \ P(h_i=1) = \sigma(\sum_{j=0}^{i-1} (W_r^{i, j<i} h_{j<i}) + b_i),
\end{align}
where $\sigma$ is the sigmoid ($\sigma(x) = 1/(1 + \exp(-x))$) function, $W_r$
is a lower-triangular square matrix, and $\vb$ is the bias vector.

For our experiments, we use conditional and approximate posterior densities that follow Bernoulli distributions:
\begin{align}
P(h_{i, l} = 1 | \vh_{l + 1}) = \sigma(W_l^{i, :} \cdot \vh_{l + 1} + b_{i, l}),
\end{align}
where $W_l$ is a weight matrix between the $l$ and $l+1$ layers. As in
\citet{gregor2013deep} with MNIST, we do not use autoregression on
the observations, $\vx$, and use a fully factorized approximate posterior.

\subsection{Variance Reduction and Choosing the AIR Objective}

\begin{table}[t]
    \begin{minipage}[l]{0.5\textwidth}
		\includegraphics[scale=0.57]{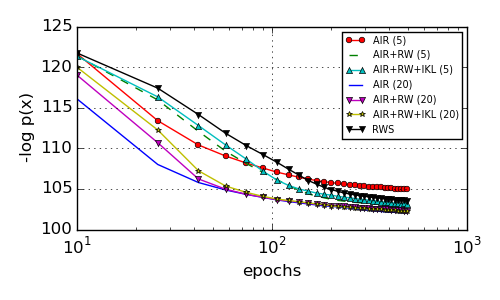}
	    
   \end{minipage}
   \begin{minipage}[r]{0.5\textwidth}
		\includegraphics[scale=0.57]{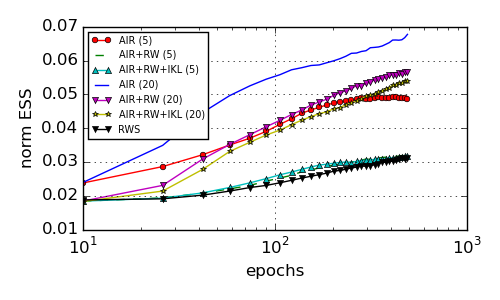}
   \end{minipage}
   \captionof{figure}{ \small
       The log-likelihood (left) and normalized effective sample size (right) with epochs in log-scale on
       the training set for AIR with $5$ and $20$ refinement steps (vanilla
       AIR), reweighted AIR with $5$ and $20$ refinement steps, reweighted AIR
       with inclusive KL objective and $5$ or $20$ refinement steps, and
       reweighted wake-sleep (RWS), all with a single stochastic latent layer.  All models were evaluated with $100$
       posterior samples, their respective number of refinement steps for the
       effective sample size (ESS), and with $20$ refinement steps of AIR for
       the log-likelihood. Despite longer wall-clock time per epoch, 
   }
   \label{fig:LLESS}
\end{table}

The effective sample size (ESS) in Eq.~\eqref{eq:ess} is a good indicator of the
variance of gradient estimate. In Fig.~\ref{fig:LLESS} (right), we observe that
the ESS improves as we take more AIR steps when training a deep belief network
(AIR(5) vs AIR(20)). When the approximate posterior is {\it not} refined (RWS),
the ESS stays low throughout training, eventually resulting in a worse model. This
improved ESS reveals itself as faster convergence in terms of the exact
log-likelihood in the left panel of Fig.~\ref{fig:LLESS} (see the progress of
each curve until 100 epochs. See also supplementary materials for wall-clock time.)

This faster convergence does not guarantee a good final log-likelihood, as the
latter depends on the tightness of the lowerbound rather than the variance of
its estimate. This is most apparent when comparing AIR(5), AIR+RW(5) and
AIR+RW+IKL(5). AIR(5) has a low variance (high ESS) but computes the gradient of
a looser lowerbound from Eq.~\eqref{eq:grad1}, while the other two compute the
gradient of a tighter lowerbound from Eq.~\eqref{eq:gradK}. This results in
AIR(5) converging faster than the other two, while the final log-likelihood
estimates are better for the other two.

We however observe that the final log-likelihood estimates are comparable across
all three variants (AIR, AIR+RW and AIR+RW+IKL) when a sufficient number of AIR
steps are taken so that $\LL_1$ is sufficiently tight. When 20 steps were taken,
we observe that the AIR(20) converges faster as well as achieves a better
log-likelihood compared to AIR+RW(20) and AIR+RW+IKL(20). Based on these
observations, we use vanilla AIR (subsequently just ``AIR'') in our following
experiments.

\subsection{Training and Density Estimation}
We evaluate AIR for training SBNs with one, two, and three layers of $200$
hidden units and DARN with $200$ and $500$ hidden units, comparing against our
implementation of RWS.  All models were tested using $100,000$ posterior samples
to estimate the lowerbounds and average test log-probabilities. 

When training SBNs with AIR and RWS, we used a completely deterministic network
for the approximate posterior.  For example, for a 2-layer SBN, the approximate
posterior factors into the approximate posteriors for the top and the bottom
hidden layers, and the initial variational parameters of the top layer, $\vmu^{(2)}_0$ are
a function of the initial variational parameters of the first layer, $\vmu^{(1)}_0$:
\begin{align}
q_0(\vh_1, \vh_2 | \vx) = q_0(\vh_1 | \vx; \vmu^{(1)}_0) q(\vh_2 | \vx; \vmu^{(2)}_0); \quad
\vmu^{(1)}_0 = f_1(\vx; \PS_1); \quad
\vmu^{(2)}_0 = f_2(\vmu^{(1)}_0; \PS_2).
\end{align}
For DARN, we trained two different configurations on MNIST: one with $500$
stochastic units and an additional hyperbolic tangent deterministic layer with
$500$ units in both the generative and recognition networks, and another with
$200$ stochastic units with a $500$ hyperbolic tangent deterministic layer in
the generative network only.  We used DARN with $200$ units
with the Caltech-101 silhouettes dataset.

\begin{table*}[t]
\caption{\small 
    Results for adaptive importance sampling iterative refinement (AIR),
    reweighted wake-sleep (RWS), and RWS with refinement with AIR at test (RWS+)
    for a variety of model configurations.  Additional sigmoid belief networks
    (SBNs) trained with neural variational inference and learning (NVIL) from
    \textdagger \citet{mnih2014neural} and variational inference for Monte Carlo
    objectives (VIMCO) from \textsection \citet{mnih2016variational}.  AIR is
    trained with $20$ inference steps and adaptive samples ($K=20, M=20$) in
    training (*3 layer SBN was trained with $50$ steps with a inference rate of
    $0.05$).  NVIL DARN results are from fDARN and VIMCO was trained using $50$
    posterior samples (as opposed to $20$ with AIR and RWS).
}
\label{table:binary}

\small
\begin{tabular}{| l || c | c  | c  || c | c || c | c | c |} 
\hline
\multirow{2}{*}{Model} & \multicolumn{5}{c||}{\bf MNIST} & \multicolumn{3}{c|}{\bf Caltech-101 Silhouettes} \\
\cline{2-9} 
\cline{2-9}
&RWS	&RWS+	&AIR	 & NVIL\textdagger	&VIMCO\textsection & RWS    &RWS+   &AIR \\	
\hline
SBN 200			&102.51	& 102.00	& {\bf 100.92}	& 113.1	& -- & 121.38& 118.63 &{\bf 116.61} \\
SBN 200-200 		&93.82	& {\bf 92.83}	& {\bf 92.90} 	&99.8	& -- & 112.86 & 107.20 & {\bf 106.94} \\
SBN 200-200-200 	&92.00	& {\bf 91.02} 	& $\text{92.56}^{*}$	&96.7	& {\bf 90.9}\textsection & 110.57& 104.54 & {\bf 104.36} \\
\hline
DARN 200 		&86.91 	&86.21	&{\bf 85.89}	&92.5\textdagger 	& -- & 113.69    & {\bf 109.73} & {\bf 109.76} \\
DARN 500 		&85.40	&{\bf 84.71}	&85.46 	&90.7\textdagger	& -- & -- & --  & --\\
\hline
\end{tabular}

%\begin{tabular}{| m{8em} | m{1cm}| m{1cm} | m{1cm} || m{1cm}| m{1.2cm} |} 
%\hline
%& \multicolumn{5}{c}{\bf MNIST} \vline\\
%\hline\hline
%Model &RWS	&RWS+	&AIR	 & NVIL\textdagger	&VIMCO\textsection\\	
%\hline
%SBN 200			&102.51	& 102.00	& {\bf 100.92}	& 113.1	& \\
%SBN 200-200 		&93.82	& {\bf 92.83}	& {\bf 92.90} 	&99.8	& \\
%SBN 200-200-200 	&92.00	& {\bf 91.02} 	& $\text{92.56}^{*}$	&96.7	& {\bf 90.9}\textsection \\
%\hline
%DARN 200 		&86.91 	&86.21	&{\bf 85.89}	&92.5\textdagger 	&\\
%DARN 500 		&85.40	&{\bf 84.71}	&85.46 	&90.7\textdagger	&\\
%\hline
%\end{tabular}
%\begin{tabular}{ | m{8em} | m{1cm}| m{1cm} | m{1cm} |} 
%\hline
%& \multicolumn{3}{c}{\bf Caltech-101 Silhouettes} \vline\\
%\hline\hline
%Model					&RWS	&RWS+	&AIR	 \\		
%\hline
% SBN 200					& 121.38& 118.63 &{\bf 116.61} \\
% SBN 200-200 				& 112.86 & 107.20 & {\bf 106.94}\\
%SBN 200-200-200			& 110.57& 104.54 & {\bf 104.36} \\
%\hline
%DARN 200 					& 113.69	& {\bf 109.73} & {\bf 109.76}\\
%\hline
%\end{tabular}					
\end{table*}
The results of our experiments with the MNIST and Caltech-101 silhouettes
datasets trained with AIR, RWS, and RWS refined at test with AIR (RWS+) are in
Table \ref{table:binary}.  Refinement at test (RWS+) always improves the results
for RWS.  As our unrefined results are comparable to those found in
\citet{bornschein2014reweighted}, the improved results indicate many evaluations
of Helmholtz machines in the literature could benefit from refinement with AIR
to improve evaluation accuracy.
%along with various other methods for training SBNs and other generative models for reference. 
For most model configurations, AIR and RWS perform comparably, though RWS
appears to do better in the average test log-probability estimates for some
configurations of MNIST. RWS+ performs comparably with variational inference for
Monte Carlo objectives \citep[VIMCO,][]{mnih2016variational}, despite the
reported VIMCO results relying on more posterior samples in training. Finally, AIR results approach SOTA with Caltech-101 silhouettes with 3-layer SBNs
against neural autoregressive distribution estimator
\citep[NADE,][]{bornschein2014reweighted}.

We also tested our log-probability estimates against the exact log-probability
(by marginalizing over the joint) of smaller single-layer SBNs with $20$
stochastic units. The exact log-probability was $-127.474$ and our estimate with
the unrefined approximate was $-127.51$ and $-127.48$ with $100$ refinement
steps. Overall, this result is consistent with those of Table
\ref{table:binary}, that iterative refinement improves the accuracy of
log-probability estimates.

\begin{figure*}
\centering
\includegraphics[width=\textwidth]{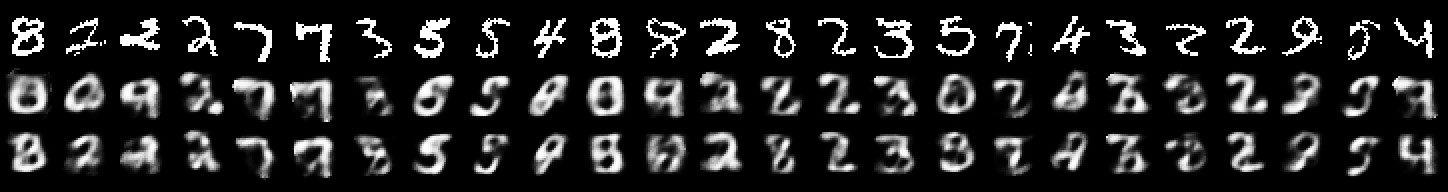}
\includegraphics[width=\textwidth]{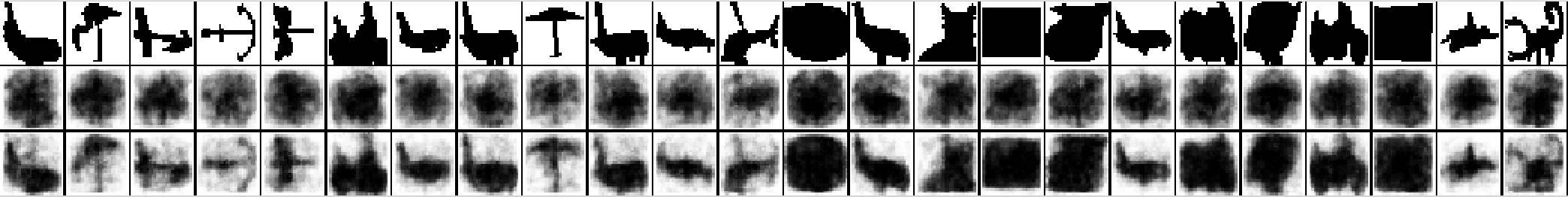}
\caption{\small Top: Average reconstructions, $1/ N \sum_{n=1}^N p(\vx | \vh^{(n)})$, for $\vh^{(n)}$ sampled from the output of the recognition network, $q_0(\vh | \vx)$ (middle row) against those sampled from the refined posterior, $q_T(\vh | \vx)$ (bottom row) for $T=20$ with a model trained on MNIST. 
Top row is ground truth. 
%We used a smaller model with $20$ latent variables to demonstrate that refinement can help even simple models fit the data well.
 Among the digits whose reconstruction changes the most, many changes correctly reveal the identity of the digit.
Bottom: Average reconstructions for a single-layer model with $200$ trained on Caltech-101 silhouettes. 
Instead of using the posterior from the recognition network, we derived a simpler version, setting $80\%$ of the variational parameters from the recognition network to $0.5$, then applied iterative refinement. 
}
\label{fig:refinement}
\end{figure*}

\subsection{Posterior Improvement}
In order to visualize the improvements due to refinement and to demonstrate AIR
as a general means of improvement for directed models at test, we generate $N$
samples from the approximate posterior without ($\vh \sim q_0(\vh | \vx; \PS)$)
and with refinement ($\vh \sim q_T(\vh | \vx)$), from a single-layer SBN with
$20$ stochastic units originally trained with RWS.  We then use the samples from
the approximate posterior to compute the expected conditional probability or
average reconstruction: $\frac{1}{N}\sum_{n=1}^N p(\vx | \vh^{(n)})$.  We used a
restricted model with a lower number of stochastic units to demonstrate that
refinement also works well with simple models, where the recognition network is
more likely to ``average'' over latent configurations, giving a misleading
evaluation of the model's generative capability. 

We also refine the approximate posterior of a simplified version of the
recognition network of a single-layer SBN with $200$ units trained with RWS. We
simplified the approximate posterior by first computing $\vmu_0 =  f(\vx; \PS)$,
then randomly setting $80\%$ of the variational parameters to $0.5$.

Fig.~\ref{fig:refinement} shows improvement from refinement for $25$ digits from
the MNIST test dataset, where the samples chosen were those of which the
expected reconstruction error of the original test sample was the most improved.
The digits generated from the refined posterior are of higher quality, and in
many cases the correct digit class is revealed. This shows that, in many cases
where the recognition network indicates that the generative model cannot model a
test sample correctly, refinement can more accurately reveal the model's
capacity. With the simplified approximate posterior, refinement is able to
retrieve most of the shape of images from the Caltech-101 silhouettes, despite
only starting with $20\%$ of the original parameters from the recognition
network. This indicates that the work of inference need not all be done via a
complex recognition network: iterative refinement can be used to aid in
inference with a relatively simple approximate posterior.

\section{Conclusion}
We have introduced iterative refinement for variational inference (IRVI), a
simple, yet effective and flexible approach for training and evaluating directed
belief networks that works by improving the approximate posterior from a
recognition network.  We demonstrated IRVI using adaptive importance refinement
(AIR), which uses importance sampling at each iterative step, and showed that
AIR can be used to provide low-variance gradients to efficiently train deep
directed graphical models.  AIR can also be used to more accurately reveal the
generative model's capacity, which is evident when the approximate posterior is
of poor quality.  The improved approximate posterior provided by AIR shows an
increased effective samples size, which is a consequence of a better
approximation of the true posterior and improves the accuracy of the test
log-probability estimates.

\section{Acknowledgements}
This work was supported by Microsoft Research to RDH (internship under NJ); NIH P20GM103472, R01 grant REB020407, and NSF grant 1539067 to VDC; and ONR grant N000141512791 and ADeLAIDE grant FA8750-16C-0130-001 to RS.
KC was supported in part by Facebook, Google (Google Faculty Award 2016) and NVidia (GPU Center of Excellence 2015-2016), and RDH was supported in part by PIBBS.

\small
\bibliography{irvi}
\bibliographystyle{plainnat}

\section{Supplementary material}
\subsection{Continuous Variables}
With variational autoencoders (VAE), the back-propagated gradient of the lowerbound with respect to the approximate posterior is composed of individual gradients for each factor, $\mu_i$ that can be applied simultaneously. 
Applying the gradient directly to the variational parameters, $\vmu$, without back-propagating to the recognition network parameters, $\PS$, yields a simple iterative refinement operator:
\begin{align}
\label{eq:GDIR}
\vmu_{t + 1} = g(\vmu_t, \vx, \gamma) = \vmu_t + \gamma \nabla_{\vmu} \LL_1(\vmu, \vx, \vepsilon),
\end{align}
where $\gamma$ is the inference rate hyperparameter and $\vepsilon$ is auxiliary noise used in the re-parameterization.

This gradient-descent iterative refinement (GDIR) is very straightforward with continuous latent variables as with VAE.
However, GDIR with discrete units suffers the same shortcomings as when passing the gradients directly,  so a better transition operator is needed (AIR).

In the limit of $T=0$, we do not arrive at VAE, as the gradients are never passed through the approximate posterior during learning. 
However, as the complete computational graph involves a series of differentiable variables, $\vmu_t$, in addition to auxiliary noise, it is possible to pass gradients through GDIR to the recognition network parameters, $\PS$, during learning, though we do not here.

For continuous latent variables, we used the same network structure as in \citep{kingma2013auto, icml2015_salimans15}.
Results for GDIR are presented in Table \ref{table:continuous} for the MNIST dataset, and included for comparison are methods for learning non-factorial latent distributions for Gaussian variables and the corresponding result for VAE, the baseline.
\begin{table}[t]
    \caption{Lowerbounds and NLL for various continuous latent variable models and training algorithms along with the corresponding VAE estimates. We use $200$ latent Gaussian variables. \textdagger From \citet{icml2015_salimans15}. \textsection From \citet{rezende2015variational}. \textdaggerdbl From \citet{burda2015importance}.}
    \label{table:continuous}
        \begin{tabular}{ | m{9em} | m{1.6cm} | m{1.6cm} | } 
            \hline
            Model & $\le$ -log p(x) & $\approx$ -log p(x) \\ 
            \hline
            \hline
            VAE & 94.48 & 89.31\\
            VAE (w/ refinement) & 90.57 & 88.53 \\
            $\text{GDIR}_{50, 20}$ & 90.60 & 88.54 \\
            \hline
            \hline
            VAE\textdagger & 94.18 & 88.95\\
            $\text{HVI}_{1}$\textdagger & 91.70 & 88.08 \\ 
            $\text{HVI}_{8}$\textdagger & 88.30 & 85.51 \\ 
            \hline
            VAE \textsection & 89.9 &\\
            $\text{DLGM+NF}_{80}$ \textsection & {\bf 85.1} &\\
            \hline
            VAE\textdaggerdbl & & 86.35 \\
            IWAE ($K=50$)\textdaggerdbl & & {\bf 84.78} \\ 
            \hline
        \end{tabular}

\end{table}

Though GDIR can improve the posterior in VAE, our results show that VAE is at an upper-bound for learning with a factorized posterior on the MNIST dataset. Further improvements on this dataset must be made by using a non-factorized posterior (re-weighting or sequential Monte Carlo with importance weighting). GDIR may still also provide improvement for training models with other datasets, and we leave this for future work.

\section{Refinement of the lowerbound and effective sample size}
\begin{table}[t]
    \begin{minipage}[l]{0.5\textwidth}
		\includegraphics[scale=0.57]{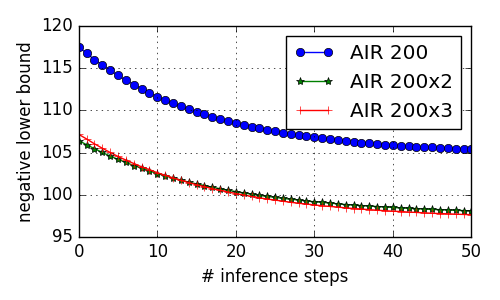}
	    
   \end{minipage}
   \begin{minipage}[r]{0.5\textwidth}
		\includegraphics[scale=0.57]{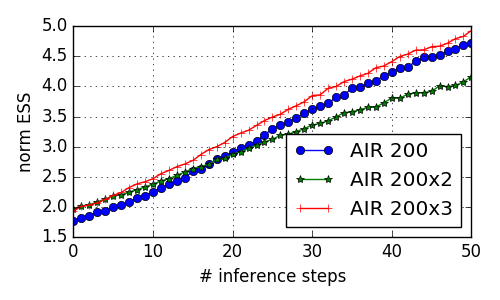}
   \end{minipage}
   \captionof{figure}{ 
   \small
       The variational lowerbound (left) and normalized effective sample size (ESS, right) the test set as the posterior is refined from the initial posterior provided by the recognition network. 
       Models were trained with AIR with $20$ refinement steps and one (AIR 200), two (AIR 200x2), and three (AIR 200x3) hidden layers.
       Refinement shows clear improves of both the variational lowerbound and effective sample size.
   }
   \label{fig:refine}
\end{table}

Iterative refinement via adaptive inference refinement (AIR) improves the variational lowerbound and effective sample size (ESS) of the approximate posterior.
To show this, we trained models with one, two, and three hidden layers with $200$ binary units trained using AIR with $20$ inference steps on the MNIST dataset for $500$ epochs.
Taking the initial approximate posterior from each model, we refined the posterior up to $50$ steps (Figure \ref{fig:refine}), evaluating the lowerbound and ESS using $100$ posterior samples.
Refinement improves the posterior from models trained on AIR well beyond the number of steps used in training.

\section{Updates and wall-clock times}
\begin{table}[t]
    \begin{minipage}[l]{0.5\textwidth}
		\includegraphics[scale=0.57]{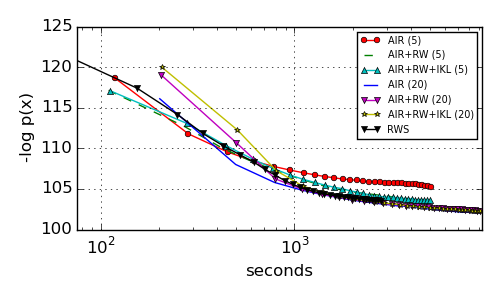}
	    
   \end{minipage}
   \begin{minipage}[r]{0.5\textwidth}
		\includegraphics[scale=0.57]{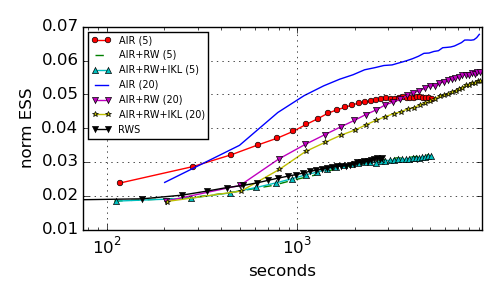}
   \end{minipage}
   \captionof{figure}{ \small
       The log-likelihood (left) and normalized effective sample size (right) with epochs in log-scale on
       the training set for AIR with $5$ and $20$ refinement steps (vanilla
       AIR), reweighted AIR with $5$ and $20$ refinement steps, reweighted AIR
       with inclusive KL objective and $5$ or $20$ refinement steps, and
       reweighted wake-sleep (RWS).  Despite longer wall-clock time per epoch, AIR converges to lower log-likelihoods and effective sample size (ESS) than RWS.
   }
   \label{fig:LLESS}
\end{table}

\begin{table}[t]
    \begin{minipage}[l]{0.5\textwidth}
		\includegraphics[scale=0.57]{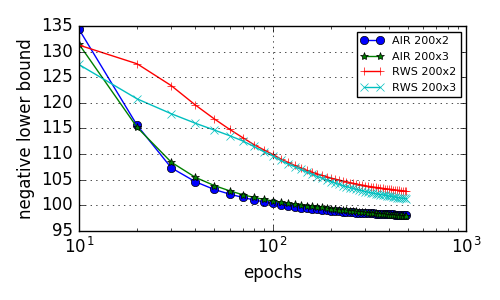}
	    
   \end{minipage}
   \begin{minipage}[r]{0.5\textwidth}
		\includegraphics[scale=0.57]{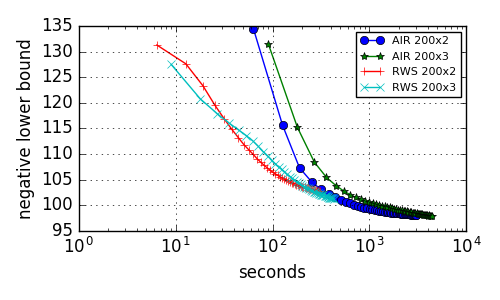}
   \end{minipage}
   \begin{minipage}[l]{0.5\textwidth}
		\includegraphics[scale=0.57]{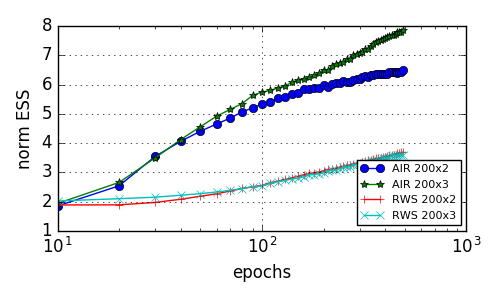}
	    
   \end{minipage}
   \begin{minipage}[r]{0.5\textwidth}
		\includegraphics[scale=0.57]{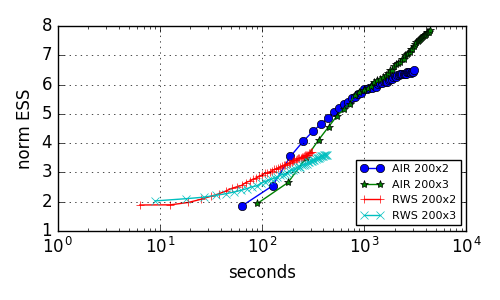}
   \end{minipage}
   \captionof{figure}{ \small
       Negative lowerbound and effective samples size (ESS) across updates (epochs) and wall-clock time (seconds) for two and three layer sigmoid belief networks trained with adaptive iterative refinement (AIR) and reweighted wake-sleep (RWS).
       AIR was trained with $20$ refinement steps with a damping rate of $\gamma = 0.9$.
       Each model was trained for $500$ epochs and evaluated on the training dataset using $100$ posterior samples.
       AIR takes less updates to reach equivalent variational lowerbound and ESS than RWS.
       While RWS can reach a higher lowerbound at earlier wall-clock times, AIR and RWS appear to converge to the same value, and AIR reaches much higher ESS.
   }
   \label{fig:lowerbounds}
\end{table}

Adaptive iterative refinement (AIR) and reweighted wake-sleep~\citep[RWS,][]{bornschein2014reweighted} have competing convergence wall-clock times, while AIR outperforms on updates (Figures~\ref{fig:LLESS} and \ref{fig:lowerbounds}).
AIR converges to a higher lowerbound and with far fewer updates than RWS, though RWS converges sooner to a similar value as AIR does later in training time.
AIR outperforms RWS in ESS in both wall-clock time and updates.
For a more accurate comparison, RWS may need to be trained at wall-clock times equal to that afforded to AIR.
However, these results support the conclusion that AIR converges to similar values as RWS in less updates but similar wall-clock times. 

\section{Bidirectional Helmholtz machines and AIR}
As an alternative to the variational lowerbound, a lowerbound can be formulated from the geometric mean of the joint generative and approximate posterior models:
\begin{align}
p^{\star}(\vx, \vh) = \frac{1}{Z}\sqrt{p(\vx, \vh) q(\vx, \vh)}.
\end{align}
In this procedure, known as bidirectional Helmholtz machines~\citep{bornschein2015bidirectional}, the lowerbound, which minimizes the Bhattacharyya distance ($D_B(p, q) = -\log \sum_y \sqrt{p(y) q(y)}$), yields estimates of the likelihood, $p^{\star}(\vx)$, with importance weights,
\begin{align}
w^{(k)} = \sqrt{\frac{p(\vx, \vh^{(k)})}{q(\vh^{(k)} | \vx)}}.
\label{eq:bidir}
\end{align}

Similar to with the variational lowerbound, we can refine the approximate posterior to maximize this lowerbound by simply replacing the weights in Equation~\ref{eq:bidir}.

We performed similar experiments to those as the experiments on wall-clock times above, using only a three layer SBN trained for $500$ epochs with the equivalent AIR and BiHM procedures using the bidirectional lowerbound importance weights.
We evaluated these models using $10000$ posterior samples on the test dataset and evaluated BiHM with (BiHM+) and without refinement.

Our results show similar negative log likelihoods for AIR ($92.40$ nats), BiHM ($93.30$ nats), and BiHM+ ($92.90$ nats), though AIR slightly outperforms BiHM+, and BiHM+ slightly outperforms BiHM.
Further optimization is necessary for a better comparison to our experiments with the variational lowerbound.
However these observations are consistent with those from our original experiments: AIR can be used to improve the posterior both in training and when evaluating models, regardless of how they were trained.
Furthermore, AIR is compatible with optimizations based on alternative lowerbounds, broadening the scope in which AIR is applicable.

\end{document}